\documentclass[11pt,a4paper]{article}
\usepackage[hyperref]{acl2019}
\usepackage{times}
\usepackage{latexsym}
\usepackage{lipsum}
\usepackage{todonotes}
\usepackage{booktabs}
\usepackage{microtype}
\usepackage[utf8]{inputenc}

\usepackage{floatrow}
\usepackage{multirow}
\usepackage{arydshln}
\usepackage{calc}

\newcommand{\textover}[3][l]{%
 \makebox[\widthof{#3}][#1]{#2}%
 }
\usepackage{url}

\aclfinalcopy 
\def\aclpaperid{682} 

\newenvironment{itemizesquish}
{\begin{list}{\labelitemi}
{\setlength{\itemsep}{0em}\setlength{\labelwidth}
{0.5em}\setlength{\leftmargin}
{\labelwidth}\addtolength{\leftmargin}
{\labelsep}}}{\end{list}}

\title{A Simple and Effective Approach to
  Automatic Post-Editing\\with Transfer Learning}

\author{
  Gon\c{c}alo M. Correia \\
  Instituto de Telecomunica\c{c}\~oes \\
  Lisbon, Portugal \\
  \texttt{goncalo.correia@lx.it.pt}
  \And
  Andr\'{e} F.~T. Martins \\
  Instituto de Telecomunica\c{c}\~oes \& Unbabel \\
  Lisbon, Portugal\\
  \texttt{andre.martins@unbabel.com}
  }

\date{}

\begin{document}
\maketitle
\begin{abstract}

  Automatic post-editing (APE) seeks to automatically refine the
  output of a black-box machine translation (MT) system through human
  post-edits. APE systems are usually trained by complementing human
  post-edited data with large, artificial data generated through
  back-translations, a time-consuming process often no easier than
  training a MT system from scratch. In this paper, we propose an
  alternative where we fine-tune pre-trained BERT models on both the
  encoder and decoder of an APE system, exploring several parameter
  sharing strategies. By only training on a dataset of 23K sentences
  for 3 hours on a single GPU we obtain results that are competitive
  with systems that were trained on 5M artificial sentences. When we
  add this artificial data, our method obtains state-of-the-art
  results.

\end{abstract}

\section{Introduction}

The goal of {\bf automatic
post-editing}~\citep[APE;][]{simard2007rule} is to automatically
correct the mistakes produced by a black-box machine translation (MT)
system. APE is particularly appealing for rapidly customizing MT,
avoiding to train new systems from scratch. Interfaces where human
translators can post-edit and improve the quality of MT
sentences~\citep{Alabau2014,Federico2014,Denkowski2015,Hokamp2018}
are a common data source for APE models, since they provide {\bf
triplets} of {\it source} sentences ({\tt src}), {\it machine
translation} outputs ({\tt mt}), and {\it human post-edits} ({\tt
pe}).

Unfortunately, human post-edits are typically scarce. Existing APE
systems circumvent this by generating {\bf artificial
triplets}~\citep{junczys2016log, negri2018escape}. However, this
requires access to a high quality MT system, similar to (or better
than) the one used in the black-box MT itself. This spoils the
motivation of APE as an alternative to large-scale MT training in the
first place: the time to train MT systems in order to extract these
artificial triplets, combined with the time to train an APE system on
the resulting large dataset, may well exceed the time to train a MT
system from scratch.

Meanwhile, there have been many successes of {\bf transfer learning}
for NLP: models such as CoVe~\citep{mccann2017learned},
ELMo~\citep{peters2018deep}, OpenAI GPT~\citep{radford2018improving},
ULMFiT~\citep{howard2018universal}, and BERT~\citep{devlin2018bert}
obtain powerful representations by training large-scale language
models and use them to improve performance in many
sentence-level and word-level tasks. However, a language generation
task such as APE presents additional challenges.

In this paper, we build upon the successes above and show that {\bf
transfer learning is an effective and time-efficient strategy for
APE}, using a pre-trained BERT model. This is an appealing strategy
in practice: while large language models like BERT are expensive to
train, this step is only done once and covers many languages,
reducing engineering efforts substantially. This is in contrast with
the computational and time resources that creating artificial
triplets for APE needs---these triplets need to be created separately
for every language pair that one wishes to train an APE system for.

Current APE systems struggle to overcome the MT baseline without
additional data. This baseline corresponds to leaving the MT
uncorrected (``do-nothing'' baseline).\footnote{If an APE system has
worse performance than this baseline, it is pointless to use it.}
With only the small shared task dataset (23K triplets), our proposed
strategy outperforms this baseline by $-$4.9 TER and $+$7.4 BLEU in
the English-German WMT 2018 APE shared task, with 3 hours of training
on a single GPU. Adding the artificial eSCAPE
dataset~\citep{negri2018escape} leads to a performance of 17.15 TER,
a new state of the art.

Our main contributions are the following:
\begin{itemizesquish}
  \item We combine the ability of BERT to
  handle {\bf sentence pair inputs} together with its pre-trained
  multilingual model, to use both the {\tt src} and {\tt mt} in a
  {\bf cross-lingual encoder}, that takes a multilingual
  sentence pair as input.
  \item We show how pre-trained BERT models can also be used and
  fine-tuned as the {\bf decoder} in a language generation task.
  \item We make a thorough empirical evaluation of different ways of
  coupling BERT models in an APE system, comparing different options
  of parameter sharing, initialization, and fine-tuning.
\end{itemizesquish}

\section{Automatic Post-Editing with BERT}\label{sec:ape_bert}

\subsection{Automatic Post-Editing}

APE~\citep{simard2007rule} is inspired by human post-editing, in
which a translator corrects mistakes made by an MT system. APE
systems are trained from triplets ({\tt src}, {\tt mt}, {\tt pe}),
containing respectively the source sentence, the machine-translated
sentence, and its post-edited version.

\paragraph{Artificial triplets.}
Since there is little data available (e.g WMT 2018 APE shared task
has 23K triplets), most research has focused on creating artificial
triplets to achieve the scale that is needed for powerful
sequence-to-sequence models to outperform the MT baseline, either
from ``round-trip'' translations~\citep{junczys2016log} or starting
from parallel data, as in the eSCAPE corpus of
\newcite{negri2018escape}, which contains 8M synthetic triplets.

\paragraph{Dual-Source Transformer.}
The current state of the art in APE uses a
Transformer~\citep{vaswani2017attention} with {\bf two encoders}, for
the {\tt src} and {\tt mt}, and {\bf one decoder}, for {\tt
pe}~\citep{junczys2018ms, tebbifakhr2018multi}. When concatenating
human post-edited data and artificial triplets, these systems greatly
improve the MT baseline. However, little successes are known using
the shared task training data only.

By contrast, with transfer learning, our work outperforms this
baseline considerably, even without any auxiliary synthetic dataset;
and, as shown in \S\ref{sec:experiments}, it achieves
state-of-the-art results by combining it with the aforementioned
artificial datasets.

\begin{figure}[htbp]
  \centering
  \includegraphics[width=\columnwidth]{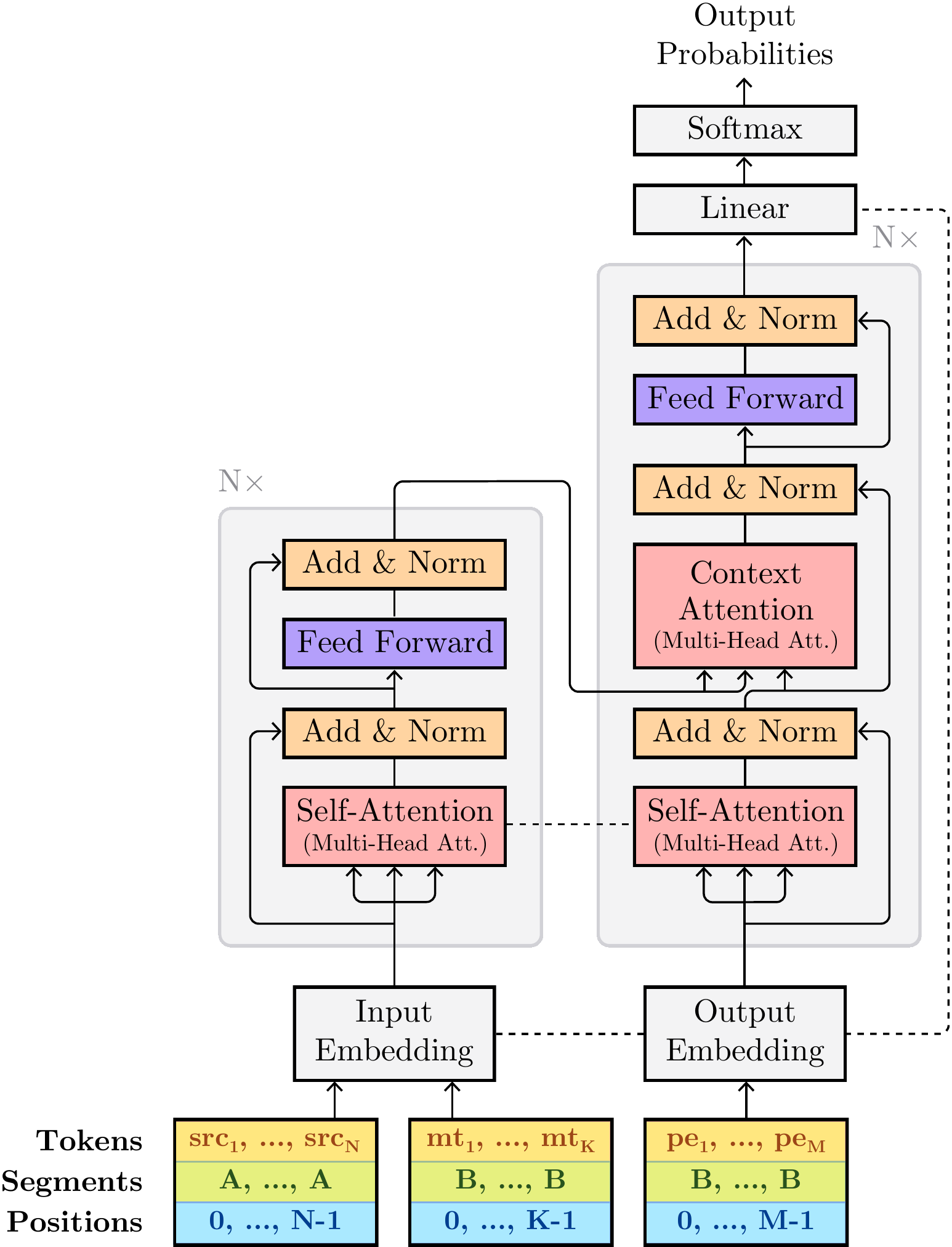}
  \caption{\textbf{Dual-Source BERT.}
  Dashed lines show shared parameters in our best configuration.}
  \label{fig:transformer_diagram}
\end{figure}

\subsection{BERT as a Cross-Lingual Encoder}

Our transfer learning approach is based on the Bidirectional Encoder
Representations from Transformers~\citep[BERT;][]{devlin2018bert}.
This model obtains deep bidirectional representations by training a
Transformer~\citep{vaswani2017attention} with a large-scale dataset
in a masked language modeling task where the objective is to predict
missing words in a sentence. We use the BERT\textsubscript{BASE}
model, which is composed of $L=12$ self-attention layers, hidden size
$H=768$, $A=12$ attention heads, and feed-forward inner layer size
$F=3072$. In addition to the word and learned position embeddings,
BERT also has {\bf segment embeddings} to differentiate between a
segment A and a segment B---this is useful for tasks such as natural
language inference, which involve two sentences. In the case of APE,
there is also a pair of input sentences ({\tt src}, {\tt mt}) which
are in different languages. Since one of the released BERT models was
jointly pre-trained on 104 languages,\footnote{
\url{https://github.com/google-research/bert/blob/master/multilingual.md}}
we use this multilingual BERT pre-trained model to encode the
bilingual input pair of APE.

Therefore, the whole encoder of our APE model is the multilingual
BERT: we encode both {\tt src} and {\tt mt} in
the same encoder and use the segment embeddings to differentiate
between languages (Figure~\ref{fig:transformer_diagram}). We reset
positional embeddings when the {\tt mt} starts, since it is not a
continuation of {\tt src}.

\subsection{BERT as a Decoder}\label{sec:ape_bert_decoder}

Prior work has incorporated pre-trained models in {\it encoders}, but
not as {\bf decoders} of sequence-to-sequence models. Doing so
requires a strategy for generating fluently from the pre-trained
model. Note that the bidirectionality of BERT is lost, since the
model cannot look at words that have not been generated yet, and it
is an open question how to learn decoder-specific blocks (e.g.
context attention), which are absent in the pre-trained model.

One of our key contributions is to use BERT in the decoder by
experimenting different strategies for initializing and sharing the
self and context attention layers and the positionwise feed-forward
layers. We tie together the encoder and decoder embeddings weights
(word, position, and segment) along with the decoder output layer
(transpose of the word embedding layer). We use the same segment
embedding for the target sentence ({\tt pe}) and the second sentence
in the encoder ({\tt mt}) since they are in the same language. The
full architecture is shown in Figure~\ref{fig:transformer_diagram}.
We experiment with the following strategies for coupling BERT
pre-trained models in the decoder:
\begin{itemizesquish}
    \item \textbf{Transformer.} A Transformer decoder as described in
    \newcite{vaswani2017attention} without any shared parameters,
    with the BERT\textsubscript{BASE} dimensions and randomly
    initialized weights.
    \item \textbf{Pre-trained BERT.} This initializes the decoder with
    the pre-trained BERT model. The only component initialized randomly
    is the context attention (CA) layer, which is absent in BERT. Unlike
    in the original BERT model---which only encodes sentences---a mask in
    the self-attention is required to prevent the model from looking to
    subsequent tokens in the target sentence.
    \item \textbf{ BERT initialized context attention.} Instead of a
    random initialization, we initialize the context attention layers
    with the weights of the corresponding BERT self-attention layers.
    \item \textbf{Shared self-attention}. Instead of just having the same
    initialization, the self-attentions (SA) in the encoder and decoder
    are tied during training.
    \item \textbf{Context attention shared with self-attention.} We take
    a step further and \emph{tie} the context attention and self
    attention weights---making all the attention transformation matrices
    (self and context) in the encoder and decoder tied.
    \item \textbf{Shared feed-forward.} We tie the feed-forward weights
    (FF) between the encoder and decoder.
\end{itemizesquish}

\section{Experiments} \label{sec:experiments}

We now describe our experimental results. Our models were implemented
on a fork of OpenNMT-py~\citep{klein2017opennmt} using a
Pytorch~\citep{paszke2017automatic} re-implementation of
BERT.\footnote{\url{https://github.com/huggingface/pytorch-pretrained-BERT}}
Our model's implementation is publicly
available.\footnote{\url{https://github.com/deep-spin/OpenNMT-APE}}

\paragraph{Datasets.}
We use the data from the WMT 2018 APE shared
task~\citep{Chatterjee2018} (English-German SMT), which consists of
23,000 triplets for training, 1,000 for validation, and 2,000 for
testing. In some of our experiments, we also use the eSCAPE
corpus~\citep{negri2018escape}, which comprises about 8M sentences;
when doing so, we oversample 35x the shared task data to cover $10\%$
of the final training data. We segment words with
WordPiece~\citep{wu2016google}, with the same vocabulary used in the
Multilingual BERT. At training time, we discard triplets with 200+
tokens in the combination of {\tt src} and {\tt mt} or 100+ tokens
in {\tt pe}. For evaluation, we use TER~\cite{snover2006study} and
tokenized BLEU~\cite{papineni2002bleu}.

\begin{table}[htbp]
  \small
  \centering
  \begin{tabular}{lcc}
  \toprule
   & TER$\downarrow$ & BLEU$\uparrow$ \\
  \midrule
  Transformer decoder & 20.33 & 69.31 \\
  Pre-trained BERT & 20.83 & 69.11 \\
  \hspace{1ex}\textcolor{gray}{\textit{with}}
  CA $\leftarrow$ SA & 18.91 & 71.81 \\
  \textover[r]
  {\hspace{1ex}\textcolor{gray}{\textit{and}}}{\hspace{1ex}\textit{with}}
  \textover[r]
  {SA $\leftrightarrow$}
  {CA $\leftarrow$} Encoder SA & \textbf{18.44} & \textbf{72.25} \\
  \textover[r]
  {\hspace{1ex}\textcolor{gray}{\textit{and}}}{\hspace{1ex}\textit{with}}
  \textover[r]
  {CA $\leftrightarrow$}{CA $\leftarrow$} SA & 18.75 & 71.83 \\
  \textover[r]
  {\hspace{1ex}\textcolor{gray}{\textit{and}}}{\hspace{1ex}\textit{with}}
  \textover[r]
  {FF $\leftrightarrow$}{CA $\leftarrow$} Encoder FF & 19.04 & 71.53 \\
  \bottomrule
  \end{tabular}
  \caption{
    Ablation study of decoder configurations, by gradually having more
    shared parameters between the encoder and decoder (trained without
    synthetic data). $\leftrightarrow$ denotes parameter tying and
    $\leftarrow$ an initialization.
  }
      \label{tab:ablation_smt}
\end{table}

\begin{table*}[htbp!]
  \small
  \centering
  \begin{tabular}{lccccccc}
  \toprule
   & & \multicolumn{2}{c}{test 2016}    & \multicolumn{2}{c}{test 2017} & \multicolumn{2}{c}{test 2018}  \\
  \cmidrule{3-4} \cmidrule{5-6}  \cmidrule{7-8}
  Model & Train Size & TER$\downarrow$ & BLEU$\uparrow$ & TER$\downarrow$ & BLEU$\uparrow$ & TER$\downarrow$ & BLEU$\uparrow$ \\
  \midrule
  MT baseline (Uncorrected) & & 24.76 & 62.11 & 24.48 & 62.49 & 24.24 & 62.99 \\
  \midrule
  \newcite{berard2017lig} & \multicolumn{1}{c}{23K} & 22.89 & --- & 23.08 & 65.57 & --- & --- \\
  \midrule
  \newcite{junczys2018ms} & \multicolumn{1}{c}{\multirow{2}{*}{5M}} & 18.92 & 70.86 & 19.49 & 69.72  & --- & --- \\
  \newcite{junczys2018ms}$\times 4$ & \multicolumn{1}{c}{} & 18.86 & 71.04 & 19.03 & 70.46  & --- & --- \\
  \midrule
  \newcite{tebbifakhr2018multi} & \multicolumn{1}{c}{\multirow{3}{*}{8M}} & --- & --- & --- & --- & 18.62 & 71.04 \\
  \newcite{junczys2018ms} & \multicolumn{1}{c}{}  & 17.81 & 72.79  & 18.10 & 71.72  & --- & --- \\
  \newcite{junczys2018ms}$\times 4$ & \multicolumn{1}{c}{} & 17.34 & 73.43 & 17.47 & 72.84  & 18.00 & 72.52 \\
  \midrule
  \midrule
  Dual-Source Transformer$^\dagger$
  & \multicolumn{1}{c}{\multirow{4}{*}{23K}} & 27.80 & 60.76  & 27.73 & 59.78  & 28.00 & 59.98 \\
  BERT Enc.\,+\,Transformer Dec. (\emph{Ours}) & \multicolumn{1}{c}{}  &  20.23 & 68.98 & 21.02 & 67.47  & 20.93 & 67.60 \\
  BERT Enc.\,+\,BERT Dec. (\emph{Ours}) & \multicolumn{1}{c}{}  &   18.88 & 71.61 &  19.03 &  70.66  & 19.34 & 70.41 \\
  BERT Enc.\,+\,BERT Dec. $\times 4$ (\emph{Ours}) & \multicolumn{1}{c}{}  & \textbf{18.05} & \textbf{72.39} &  \textbf{18.07} & \textbf{71.90}  & \textbf{18.91} & \textbf{70.94} \\
  \midrule
  BERT Enc.\,+\,BERT Dec. (\emph{Ours}) & \multicolumn{1}{c}{\multirow{2}{*}{8M}} & 16.91 & 74.29 & 17.26 & 73.42  & 17.71 & 72.74 \\
  BERT Enc.\,+\,BERT Dec. $\times 4$ (\emph{Ours}) & \multicolumn{1}{c}{} & \textbf{16.49} & \textbf{74.98} & \textbf{16.83} & \textbf{73.94} & \textbf{17.15} & \textbf{73.60}  \\
  \bottomrule
  \end{tabular}
  \caption{Results on the WMT 2016--18 APE shared task datasets. Our
  single models trained on the 23K dataset took only 3h20m to converge
  on a single Nvidia GeForce GTX 1080 GPU, while results for models
  trained on 8M triplets take approximately 2 days on the same GPU.
  Models marked with ``$\times 4$'' are ensembles of 4 models.
  Dual-Source Transformer$^\dagger$ is a comparable re-implementation
  of \newcite{junczys2018ms}.}
    \label{tab:results_smt}
\end{table*}

\paragraph{Training Details.}
We use Adam~\citep{kingma2014adam} with a triangular learning rate
schedule that increases linearly during the first 5,000 steps until
$5\times 10^{-5}$ and has a linear decay afterwards. When using BERT
components, we use a $\ell_2$ weight decay of $0.01$. We apply
dropout~\cite{srivastava2014dropout} with $p_{drop}=0.1$ to all
layers and use label smoothing with
$\epsilon=0.1$~\citep{pereyra2017regularizing}. For the small data
experiments, we use a batch size of 1024 tokens and save checkpoints
every 1,000 steps; when using the eSCAPE corpus, we increase this to
2048 tokens and 10,000 steps. The checkpoints are created with the
exponential moving average strategy of \newcite{junczys2018marian}
with a decay of $10^{-4}$. At test time, we select the model with
best TER on the development set, and apply beam search with a beam
size of 8 and average length penalty.

\paragraph{Initialization and Parameter Sharing.}
Table~\ref{tab:ablation_smt} compares the different decoder
strategies described in \S\ref{sec:ape_bert_decoder} on the WMT 2018
validation set. The best results were achieved by sharing the
self-attention between encoder and decoder, and by initializing (but
not sharing) the context attention with the same weights as the
self-attention. Regarding the self-attention sharing, we hypothesize
that its benefits are due to both encoder and decoder sharing a
common language in their input (in the {\tt mt} and {\tt pe}
sentence, respectively). Future work will investigate if this is
still beneficial when the source and target languages are less
similar. On the other hand, the initialization of the context
attention with BERT's self-attention weights is essential to reap the
benefits of BERT representations in the decoder---without it, using
BERT decreases performance when compared to a regular transformer
decoder. This might be due to the fact that context attention and
self-attention share the same neural block architecture (multi-head
attention) and thus the context attention benefits from the
pre-trained BERT's better weight initialization. No benefit was
observed from sharing the feed-forward weights.

\paragraph{Final Results.} 
Finally, Table~\ref{tab:results_smt} shows our results on the WMT
2016--18 test sets. The model named \emph{BERT Enc.\,+\,BERT Dec.}
corresponds to the best setting found in
Table~\ref{tab:ablation_smt}, while \emph{BERT Enc.\,+\,Transformer
Dec.} only uses BERT in the encoder. We show results for single
models and ensembles of 4 independent models.

Using the small shared task dataset only (23K triplets), our single
\emph{BERT Enc.\,+\,BERT Dec.} model surpasses the MT baseline by a
large margin ($-$4.90 TER in test 2018). The only system we are aware
to beat the MT baseline with only the shared task data is
\citet{berard2017lig}, which we also outperform ($-$4.05 TER in test
2017). With only about 3 GPU-hours and on a much smaller dataset, our
model reaches a performance that is comparable to an ensemble of the
best WMT 2018 system with an artificial dataset of 5M triplets
($+$0.02 TER in test 2016), which is much more expensive to train.
With 4$\times$ ensembling, we get competitive results with systems
trained on 8M triplets.

When adding the eSCAPE corpus (8M triplets), performance surpasses
the state of the art in all test sets. By ensembling, we improve even
further, achieving a final 17.15 TER score in test 2018 ($-$0.85 TER
than the previous state of the art).

\section{Related Work}\label{sec:related}

In their Dual-Source Transformer model, \newcite{junczys2018ms} also
found gains by tying together encoder parameters, and the embeddings
of both encoders and decoder. Our work confirms this but shows
further gains by using segment embeddings and more careful sharing
and initialization strategies. \newcite{sachan2018parameter} explore
parameter sharing between Transformer layers. However, they focus on
sharing decoder parameters in a \emph{one-to-many} multilingual MT
system. In our work, we share parameters between the encoder and the
decoder.

As stated in \S\ref{sec:experiments}, \citet{berard2017lig} also
showed improved results over the MT baseline, using
exclusively the shared task data. Their system outputs edit
operations that decide whether to insert, keep or delete tokens from
the machine translated sentence. Instead of relying on edit
operations, our approach mitigates the small amount of data with
transfer learning through BERT.

Our work makes use of the recent advances in transfer learning for
NLP~\citep{peters2018deep, howard2018universal, radford2018improving,
devlin2018bert}. Pre-training these large language models has largely
improved the state of the art of the GLUE
benchmark~\citep{wang2018glue}.
Particularly, our work uses the BERT pre-trained model and makes use
of the representations obtained not only in the encoder but also on
the decoder in a language generation task.

More closely related to our work, \newcite{lample2019cross}
pre-trained a BERT-like language model using parallel data, which
they used to initialize the encoder and decoder for supervised and
unsupervised MT systems. They also used segment embeddings (along
with word and position embeddings) to differentiate between a pair of
sentences in different languages. However, this is only used in one
of the pre-training phases of the language model (translation
language modelling) and not in the downstream task. In our work, we
use segment embeddings during the downstream task itself, which is a
perfect fit to the APE task.

\newcite{lopes2019unbabels} used our model on the harder
English-German NMT subtask to obtain better TER performance than previous
state of the art. To obtain this result, the transfer learning
capabilities of BERT were not enough and further engineering effort
was required. Particularly, a conservativeness factor was added
during beam decoding to constrain the changes the APE system
can make to the {\tt mt} output. Furthermore, the authors used a data
weighting method to augment the importance of data samples that have
lower TER. By doing this, data samples that
required less post-editing effort are assigned higher weights
during the training loop. Since the NMT system does very few errors
on this domain this data weighting is important for the APE model to
learn to do fewer corrections to the {\tt mt} output. However, their
approach required the creation of an artificial dataset to obtain a
performance that improved the MT baseline. We leave it for future work
to investigate better methods to obtain results that improve the
baseline using only real post-edited data in these smaller
APE datasets.

\section{Conclusion and Future Work}

We proposed a transfer learning approach to APE using BERT
pre-trained models and careful parameter sharing. We explored various
ways for coupling BERT in the decoder for language generation.
We found it beneficial to initialize the context
attention of the decoder with BERT's self-attention
and to tie together the parameters of the self-attention
layers between the encoder and decoder.
Using a small dataset, our results are competitive with systems
trained on a large amount of artificial data, with much faster
training. By adding artificial data, we obtain a new state of the art
in APE.

In future work, we would like to do an extensive analysis on the
capabilities of BERT and transfer learning in general for different domains and
language pairs in APE.

\section*{Acknowledgments}

This work was supported by the European Research Council (ERC StG
DeepSPIN 758969), and by the Funda\c{c}\~ao para a Ci\^encia e
Tecnologia through contracts UID/EEA/50008/2019 and
CMUPERI/TIC/0046/2014 (GoLocal). We thank the anonymous reviewers for
their feedback.

\bibliography{acl2019}
\bibliographystyle{acl_natbib}

\end{document}